\documentclass[runningheads]{llncs}
\usepackage[T1]{fontenc}
\usepackage{graphicx}
\usepackage{subcaption}
\usepackage{amssymb,amsmath,array}
\usepackage{siunitx}
\usepackage{hyperref}
\usepackage[table]{xcolor}
\usepackage{booktabs}
\usepackage{fontawesome5}

\begin{document}

\title{Do Metrics for Counterfactual Explanations Align with User Perception?}

\author{Felix Liedeker\orcidID{0009-0006-2556-9430}\faIcon{envelope} \and
Basil Ell\orcidID{0000-0002-8863-3157} \and
Philipp Cimiano\orcidID{0000-0002-4771-441X} \and
Christoph Düsing\orcidID{0000-0002-7817-9448}}

\authorrunning{F. Liedeker et al.}

\institute{Bielefeld University, Bielefeld, Germany\\
\email{\faIcon{envelope}~felix.liedeker@uni-bielefeld.de}}
\maketitle              
\begin{abstract}
Explainability is widely regarded as essential for trustworthy artificial intelligence systems. However, the metrics commonly used to evaluate counterfactual explanations are algorithmic evaluation metrics that are rarely validated against human judgments of explanation quality. This raises the question of whether such metrics meaningfully reflect user perceptions.
We address this question through an empirical study that directly compares algorithmic evaluation metrics with human judgments across three datasets. Participants rated counterfactual explanations along multiple dimensions of perceived quality, which we relate to a comprehensive set of standard counterfactual metrics. We analyze both individual relationships and the extent to which combinations of metrics can predict human assessments.
Our results show that correlations between algorithmic metrics and human ratings are generally weak and strongly dataset-dependent. Moreover, increasing the number of metrics used in predictive models does not lead to reliable improvements, indicating structural limitations in how current metrics capture criteria relevant for humans. Overall, our findings suggest that widely used counterfactual evaluation metrics fail to reflect key aspects of explanation quality as perceived by users, underscoring the need for more human-centered approaches to evaluating explainable artificial intelligence.

\end{abstract}

\section{Introduction}

Machine Learning systems are increasingly deployed in settings where transparency and user understanding play a central role. \textit{Explainable Artificial Intelligence} (XAI) seeks to address this need by providing insights into model predictions and supporting trust, accountability, and informed decision making~\cite{adadi2018peeking}. Among the many approaches proposed in recent years, \textit{counterfactual explanation} (CF) methods have emerged as a particularly influential class of methods~\cite{dusing2024integrating}. CFs describe how a model’s prediction would change under minimal modifications of an input instance, thereby offering explanations that resemble human reasoning about alternatives and actionable changes~\cite{wachter2017counterfactual}. Their contrastive, decision-oriented form aligns with how people naturally reason about ``what if'' scenarios, which has contributed to their prominence in XAI research~\cite{dusing2024integrating}.

As counterfactual methods mature, evaluating their quality becomes increasingly important. A broad range of algorithmic metrics (i.e., metrics than can be evaluated without human involvement by means of computation) has been proposed to quantify properties of CFs such as sparsity and proximity~\cite{verma2020counterfactual,Nauta_2023}. These metrics form the backbone of most comparative evaluations and are frequently used as proxies for explanation quality. In parallel, human-centered evaluations through user studies or expert judgments are widely regarded as the gold standard for assessing explanations~\cite{Doshi-Velez2017_RigorousScienceInterpretableMachine,liedeker2024empirical}. However, these two evaluation paradigms are typically applied in isolation. It therefore remains largely unclear whether commonly used automated CF metrics reflect the aspects of explanations that humans perceive as meaningful, useful, or trustworthy.

Evidence from XAI literature raises broader concerns about relying solely on automated evaluation metrics. The \textit{M4 benchmark} demonstrated that widely used faithfulness metrics for feature attribution methods correlate only weakly with one another and can produce contradictory method rankings~\cite{li2023m4}. Outside XAI, similar mismatches have been documented between automated similarity measures and human perception, for example in the evaluation of reconstructed images, where standard metrics aligned poorly with human judgments of recognizability~\cite{sun2023privacy}. These findings motivate a closer examination of whether automated metrics for CFs capture what users value in explanations.

In this work, we address this question through a systematic empirical study of the relationship between automated CF metrics and human judgments. We collect human evaluations of CFs across multiple datasets and quality dimensions, and compare these ratings to a comprehensive set of established CF metrics. Beyond analyzing individual metric--rating relationships, we investigate whether (linear and non-linear) combinations of metrics better approximate human judgments. Our results show that associations between automated metrics and human perception are generally weak and highly dataset-dependent.
Increasing the number of metrics does not improve out-of-sample performance and often degrades it, suggesting that existing metrics do not provide complementary information. Together, these findings point to a structural mismatch between what current CF metrics quantify and what users perceive as explanation quality. The main contributions of this paper are:

\begin{enumerate}
    \item We conduct a controlled user study in which participants evaluate CFs along multiple dimensions of perceived explanation quality across three widely used datasets.
    \item We compute a comprehensive set of widely used automated CF metrics for the same explanations and quantify their alignment with human ratings.
    \item We analyze whether combining multiple metrics improves the prediction of human judgments and show that increasing the number of metrics does not improve, and often degrades, predictive power.
    \item We discuss implications for the evaluation of CFs and argue for the development of metrics that are more directly grounded in human perception.
\end{enumerate}

\section{Related Work}

Evaluation of CFs has largely followed two parallel tracks: automated, metric-based assessment and human-centered evaluation. On the automated side, a wide range of metrics has been proposed to quantify properties such as validity, proximity, sparsity, plausibility, and diversity~\cite{verma2020counterfactual,Nauta_2023}. These measures are primarily motivated by computational desiderata and optimization objectives, and they dominate evaluations of CF methods. However, they are typically introduced without empirical grounding in how users perceive explanation quality~\cite{Doshi-Velez2017_RigorousScienceInterpretableMachine,adadi2018peeking}.

In contrast, human-centered research emphasizes that explanation quality is inherently psychological and context-dependent. Insights from cognitive science show that people naturally reason about alternatives to reality, but that counterfactual and causal explanations can lead to different mental models and do not always increase satisfaction or perceived usefulness~\cite{byrne2019counterfactuals}. Empirical studies of CFs report similarly nuanced findings. CFs can support understanding and richer mental models, yet they are not always preferred over simpler alternatives~\cite{warren2022features}. Other work highlights divergences between objective task performance and subjective experience: CFs may improve comprehension without necessarily increasing trust or satisfaction, or vice versa~\cite{van2021interpretable,vannostrand2024actionable,wang2021explanations}. These findings indicate that human judgments of explanation quality depend on multiple interacting factors and cannot be reduced to a single computational criterion.

A related line of work seeks to model the structure of human judgments of CFs. Domnich et al.~\cite{domnich2025predicting} analyze how multiple human-rated explanatory qualities --- such as feasibility, trust, and completeness --- predict overall user satisfaction, revealing a structured, multi-dimensional perception of explanation quality. Similarly, Kuhl et al.~\cite{kuhl2022keep} show that psychologically from a user perspective plausible CFs may differ from those that are computationally constrained to lie on the data manifold. While these approaches deepen our understanding of how people evaluate CFs, they rely on human-provided ratings as inputs and therefore do not address whether computed CF metrics can serve as proxies for human evaluation.

Beyond CFs, concerns about metric validity have emerged in XAI more broadly. The \textit{M4 benchmark} showed that widely used faithfulness metrics for feature attribution methods often correlate weakly and can lead to inconsistent rankings~\cite{li2023m4}. Similar mismatches between automated metrics and human perception have been observed in other domains, such as image reconstruction, where standard similarity measures aligned poorly with human judgments~\cite{sun2023privacy}. 

Taken together, prior work highlights (i) the central role of automated metrics in CF evaluation, (ii) the complex and multi-dimensional nature of human judgments of explanation quality, and (iii) growing evidence that automated metrics in XAI and related fields may not faithfully reflect human perception. However, systematic empirical evidence on whether widely used automated CF evaluation metrics align with human judgments of CF quality remains limited. This question is the focus of the present study.

\section{Data Acquisition and Study Design}

This section describes the datasets, CF generation, sampling strategy, user study design, and evaluation measures used in our analysis. The goal is to collect human judgments of CF quality that cover a diverse range of explanations and their properties, enabling systematic comparison with a set of automated metrics.

\subsection{Datasets and Counterfactual Generation}

To generate a diverse set of CFs while keeping the study accessible to non-expert participants, we selected three tabular classification datasets from the UCI Machine Learning Repository: \textit{Mushroom} (MUS),\footnote{\url{https://archive.ics.uci.edu/dataset/73/}} \textit{Estimation of Obesity Levels Based on Eating Habits and Physical Condition} (OBE),\footnote{\url{https://archive.ics.uci.edu/dataset/544/}} and \textit{Heart Disease} (HRT).\footnote{\url{https://archive.ics.uci.edu/dataset/45/}}
The task associated with MUS is predicting whether a mushroom is edible or poisonous based on observable attributes; OBE targets the prediction of obesity level categories from eating habits and physical condition variables; and HRT addresses the prediction of heart disease presence from clinical measurements.
These datasets vary in sample size, feature dimensionality, and label structure, covering both binary (MUS, HRT) and multi-class (OBE) prediction tasks. Table~\ref{table:dataset_overview} summarizes the dataset characteristics.
Our inclusion criteria for dataset selection were: (1) classification problems with intuitive features and labels,\footnote{For example, the Mushroom dataset includes features such as \textit{cap color: green} and \textit{odor: fishy}, which are easily understood without domain expertise.} supporting intuitive counterfactual reasoning, (2) sufficient dataset size to train stable models and produce meaningful CFs, and (3) relevance to real-world tabular decision-making scenarios. 
We excluded datasets that primarily target regression or non-classification objectives, require substantial domain expertise to interpret, are too small to support reliable modeling and explanation generation, or are so complex that they would likely overwhelm participants.

\begin{table}[t]
\centering
\caption{Overview of dataset characteristics, classification performance (using XGBoost), and number of successfully generated counterfactuals.}
\label{table:dataset_overview}
\begin{tabular}{lrrrrrr}
\toprule
Dataset & Instances & Features & Classes & F1 Score & Test Inst. & Valid CFs \\
\midrule
Mushroom (MUS)        & 8124 & 22 & 2 & 1.00 & 1625 & 755 \\
Obesity Levels (OBE) & 2111 & 16 & 7 & 0.95 & 423  & 211 \\
Heart Disease (HRT)  & 303  & 13 & 2 & 0.85 & 60   & 25 \\
\bottomrule
\end{tabular}
\end{table}

Categorical features were one-hot encoded and continuous features were min-max scaled to $[-1,1]$. Among several evaluated classifiers, XGBoost achieved the highest predictive performance (F1 scores $\ge0.85$ across datasets) and was therefore used as the base model for CF generation. After a 80:20 train-test split, CFs were generated for all instances in the test sets using the \textit{Counterfactuals Guided by Prototypes} method~\cite{Looveren2020_InterpretableCounterfactualExplanationsGuided}, implemented in the open source Python library \textit{Alibi Explain}~\cite{Klaise2021_AlibiExplainAlgorithmsExplaining}. This method generates CFs by searching for instances close to learned target-class prototypes, encouraging both proximity to the original instance and plausibility with respect to the data distribution. As is common in practice, CF generation was not always successful, especially for smaller datasets; however, the number of valid explanations was sufficient for subsequent sampling and analysis. A CF is considered valid if the generated instance is predicted by the model to belong to the desired target class, that is, if the model's prediction has flipped from the original class to the target class. It is considered invalid if the generation process fails to find such an instance, meaning the model's prediction on the generated instance remains unchanged from the original instance. The number of test set instances as well as the number of generated valid CFs are given in Table~\ref{table:dataset_overview}.

\subsection{Sampling of Counterfactual Explanations}

Given the large number of generated CFs, only a subset was included in the user study. The sampling strategy was designed to balance two competing objectives: (i) maximizing diversity such that the selected explanations reflect the structure of the full explanation space, and (ii) limiting the total number of explanations to ensure that each receives a sufficient number of human ratings.

To this end, we applied a cluster-preserving sampling procedure based on seven automated CF metrics (see Section~\ref{subsec:metrics}). 
First, k-means clustering was performed on the metric vectors to group explanations with similar quantitative properties. We then allocated samples proportionally to cluster sizes, ensuring that each cluster contributed at least one explanation. Any remaining samples were assigned based on fractional remainders so that the final allocation closely matched the underlying metric distribution. Within each cluster, explanations were drawn uniformly at random and we selected a total of 85 CFs (MUS: 30, OBE: 30, HRT: 25), which form the basis of the user study.

\subsection{User Study Procedure}

The study was conducted using \textit{Prolific}. Participants were first asked demographic questions and then introduced to CFs using an example from the simplified \textit{German Credit Risk} dataset.\footnote{\url{https://www.kaggle.com/datasets/uciml/german-credit}} Each explanation was presented as a table comparing the original instance with the CF, with changed features visually highlighted and accompanied by a brief textual description. 

The 85 CFs were organized into 20 batches of 12 explanations each (four per dataset). Each participant was randomly assigned to one batch and evaluated only the explanations in that batch. The dataset order was randomized, while same-dataset explanations were presented consecutively to reduce cognitive load. 

Participants rated each explanation along five dimensions by answering the corresponding question using a 4-point Likert scale (1=\textit{Definitely Yes}, 4=\textit{Definitely No}, plus \textit{I don’t know}):

\begin{itemize}
    \item \textbf{Perceived Accuracy}: Is the class predicted by the model accurate?
    \item \textbf{Understandability}: Is the provided explanation understandable?
    \item \textbf{Plausibility}: Is the provided explanation plausible?
    \item \textbf{Sufficiency of Detail}: Has the provided explanation sufficient detail?
    \item \textbf{User Satisfaction}: Is the provided explanation satisfying?
\end{itemize}

These dimensions follow established explanation satisfaction and trust scales and align with prior XAI user studies that emphasize \textit{understandability}, \textit{completeness}, \textit{accuracy}, and \textit{satisfaction} as core components of CF quality~\cite{abbaspour2025dynamics}.

After completing all ratings, participants answered a short post-study questionnaire including attention checks and task difficulty ratings. One participant failed the attention checks and these judgments were discarded. The final analyzed sample comprised 167 participants. Each participant received a compensation of \pounds2.75\footnote{To match the German minimum wage at the time of data collection (\texteuro12.82/hour).} and the median completion time was 15 minutes. Ages ranged from 20 to 72 ($M=40.85$, $SD=13.05$); 77.8\% held a higher education degree, and self-reported machine learning experience averaged 2.67 ($SD=0.93$) on a 1--4 scale (1=\textit{No experience}, 4=\textit{Extensive experience}).

\subsection{Power Analysis}

To assess whether the collected sample size was sufficient for our primary analyses, we conducted a power analysis for both correlation (Fisher's z-transformation) and regression (F-test, seven predictors) frameworks. Assuming a significance level of $\alpha=0.05$, our per-dataset samples ($n=25/30$ explanations) provide 80\% power to detect large effects ($r\ge0.50$) as defined by Cohen's conventions for effect sizes~\cite{cohen1988statistical}. For the predictive modeling analysis using seven metrics as predictors, our sample size ($n=25/30$ explanations) provides 80\% power to detect $R^2 \geq 0.40$; detecting $R^2 = 0.30$ and $R^2 = 0.15$ with 80\% power would require $n\approx42$ and $n\approx89$ explanations per dataset, respectively.

We consider this appropriate for an initial investigation: the central question of this study is whether automated metrics can serve as meaningful proxies for human judgments. If a single metric or a combination of metrics were to closely approximate human assessments, this would manifest as a large effect, which our study is well-powered to detect. Conversely, metrics only weakly correlated with user perception ($r < 0.30$) are arguably of limited practical utility. Our study is thus well-powered to detect practically significant effects, while acknowledging that detecting small or medium-sized effects would require larger samples.

\subsection{Rating Aggregation and Reliability}

Across all participants, we collected 2004 individual ratings, corresponding to a mean of 23.58 ($SD=2.57$) complete rating sets per explanation. To assess inter-rater reliability, we computed two-way random-effects intraclass correlation coefficients ICC(2,1) and ICC(2,k)~\cite{Shrout1979}. As expected for subjective Likert-scale judgments, individual-level agreement was low, while aggregated ratings showed higher but still modest reliability, consistent with prior XAI user studies~\cite{Tinsley1975,Wang2025}.

Internal consistency across the five rating dimensions was high (Cronbach’s $\alpha=0.88$), and a principal component analysis revealed clear unidimensionality, with the first component explaining 74.1\% of the variance. This indicates that the five dimensions capture a coherent underlying notion of perceived explanation quality. We therefore aggregate them into a single \textit{Combined Quality Score} (CQS) by averaging the five ratings per explanation, used in subsequent analyses. Table~\ref{tab:dataset_evaluation} reports the mean ratings for each dimension and the CQS.

\begin{table}[t]
\centering
\caption{Mean ratings per dataset across all dimensions and the CQS.}
\small
\begin{tabular}{lcccccc}
\toprule
& Acc. & Und. & Plaus. & Suff. & Sat. & CQS \\
\midrule
MUS      & 2.22$\pm$0.23 & 1.83$\pm$0.16 & 2.08$\pm$0.26 & 2.08$\pm$0.23 & 2.08$\pm$0.20 & 2.06$\pm$0.17 \\
HRT  & 2.25$\pm$0.27 & 1.95$\pm$0.19 & 2.24$\pm$0.23 & 2.20$\pm$0.23 & 2.29$\pm$0.24 & 2.19$\pm$0.20 \\
OBE       & 2.14$\pm$0.30 & 1.85$\pm$0.27 & 2.16$\pm$0.33 & 2.10$\pm$0.25 & 2.19$\pm$0.33 & 2.09$\pm$0.27 \\
\midrule
All          & 2.20$\pm$0.27 & 1.87$\pm$0.22 & 2.15$\pm$0.28 & 2.12$\pm$0.24 & 2.18$\pm$0.28 & 2.10$\pm$0.22 \\
\bottomrule
\end{tabular}
\label{tab:dataset_evaluation}
\end{table}

\subsection{Automated Metrics}
\label{subsec:metrics}

We compute seven widely used automated metrics for CFs as defined in related works~\cite{verma2020counterfactual,mothilal2020}: \textit{Sparsity}, \textit{Proximity}, \textit{Closeness to the training data}, \textit{Diversity}, \textit{Oracle Score}, \textit{Trust Score}, and \textit{Completeness}. These metrics capture complementary properties such as minimality, plausibility, manifold adherence, and model confidence. In the following, we briefly describe each metric:\\

\noindent\textbf{Sparsity} measures the number of features that differ between the original input $x$ and the CF $x'$:
\[
\mathrm{Sparsity}(x, x') = \sum_{i=1}^d \mathbf{1}(x_i \neq x'_i),
\]
where $d$ is the number of input features and $\mathbf{1}(\cdot)$ is the indicator function. A lower \textit{sparsity} indicates that fewer features were modified.\\

\noindent\textbf{Proximity} measures how close the generated explanation $x'$ is to the original input $x$ in feature space, typically measured using an $\ell_p$ norm, with $p=1$:
\[
    \mathrm{Proximity}_p(x,x')
    \;=\;\|x - x'\|_p
    \;=\;\Bigl(\sum_{i=1}^d \lvert x_i - x'_i\rvert^p\Bigr)^{1/p}.
\]
Thus, a smaller \textit{proximity} implies that the CF is closer to the original input.\\

\noindent\textbf{Closeness} captures how close the CF $x'$ lies to the data manifold, computed as the average distance to its $k$ nearest neighbors in the training set $\mathcal{D}_{\text{train}}$:
\[
\mathrm{Closeness}_p(x') = \frac{1}{k} \sum_{i=1}^k \|x' - x_{(i)}\|_p,
\]
where $x_{(1)}, \ldots, x_{(k)}$ denote the $k$ nearest neighbors of $x'$ in $\mathcal{D}_{\text{train}}$, ordered by distance. A lower \textit{closeness} indicates that the CF is closer to observed data. We use $p=1$ and $k=5$.\\

\noindent\textbf{Diversity} measures the heterogeneity of features modified in a CF. It is computed as the average pairwise dissimilarity across all pairs of changed features. For CFs with at least two changed features ($|C| \geq 2$), it is defined as:
\[
\mathrm{Diversity}(x, x') = 
\frac{2}{\lvert C\rvert(\lvert C\rvert - 1)} 
\sum_{\substack{i,j \in C \\ i < j}}
\bigl(1 - \mathrm{NMI}(f_i, f_j)\bigr),
\]
where $C=\{i \mid x_i \neq x'_i\}$ is the set of indices of features that differ between the original input $x$ and the CF $x'$, and $\mathrm{NMI}(f_i, f_j) \in [0,1]$ denotes the normalized mutual information between features $f_i$ and $f_j$. A higher \textit{diversity} indicates that the changed features are more independent of one another.\\

\noindent\textbf{Oracle Score} measures the plausibility of a CF by evaluating whether two independently trained classifiers agree that $x'$ belongs to the target class $c$. The score is computed as the product of the predicted class probabilities from the base model (here: XGBoost) and an oracle model (here: Random Forest):
\[
       \mathrm{OracleScore}(x', c) = \hat{P}_1(c \mid x') \cdot \hat{P}_2(c \mid x'),
\]
where $\hat{P}_1(c \mid x')$ and $\hat{P}_2(c \mid x')$ denote the probability assigned to class $c$ by the base model and oracle, respectively. A higher \textit{oracle score} indicates stronger cross-model agreement that the CF belongs the desired target class. While neither model's probabilities are explicitly calibrated, this is mitigated by the fact that our analysis (cf. Section~\ref{sec:results}) relies on rank-based correlations, which are invariant to monotone distortions of the probability values.\\

\noindent\textbf{Trust Score} quantifies the reliability of a predicted classification by comparing how much closer the instance is to its predicted class than to any other class~\cite{Jiang2018_TrustNotTrustClassifier}. For each class $c$, a KD‐tree is built from training instances of that class. The \textit{trust score} is then defined as:
\[
\mathrm{TrustScore}(x') = \frac{d_{\text{other}}}{d_{\text{pred}}}, \textrm{ where}\\
\]
\[
       d_{\text{pred}}
       =\frac{1}{k}\sum_{i=1}^k \bigl\|x' - \mathrm{NN}_i^{\hat{y}}(x')\bigr\|_p
       \quad \text{and} \quad
       d_{\text{other}}
       =\min_{c\neq \hat{y}}
       \frac{1}{k}\sum_{i=1}^k \bigl\|x' - \mathrm{NN}_i^c(x')\bigr\|_p.
\]
Here, $\hat{y}$ is the predicted class for $x'$, and $\mathrm{NN}_i^{c}(x')$ denotes the $i$-th nearest neighbor of $x'$ among training instances of class $c$. A higher \textit{trust score} indicates that $x'$ lies closer to the predicted class than to alternative classes. We use $p=1, k=3$.\\

\noindent\textbf{Completeness} measures how much of the model's feature importance is captured by the features changed in the CF. Using SHAP~\cite{lundberg2017unified} to compute feature attributions, we define:
\[
\mathrm{Completeness}_k(x, x') = \frac{\sum_{i \in C \cap T_k} \lvert\phi_i\rvert}{\sum_{j \in T_k} \lvert\phi_j\rvert},
\]
where $\phi_i$ denotes the SHAP importance of feature $i$ for the original instance $x$, $C=\{i \mid x_i \neq x'_i\}$ is the set of features changed in the CF and $T_k$ is the set of the top-$k$ features by absolute SHAP importance. A higher \textit{completeness} indicates that the CF modifies features that the model considers important for its prediction. $k=5$ is used in our calculations.

\section{Results} \label{sec:results}

We analyze the relationship between automated CF evaluation metrics and human judgments using two complementary approaches. First, we examine pairwise associations between individual metrics and human ratings to assess whether any metric consistently aligns with perceived explanation quality (Section \ref{sec:correlation}). Second, we evaluate whether combinations of metrics can predict human judgments using supervised predictive models (Section \ref{sec:predicting}).

\subsection{Metric--Rating Correlations} \label{sec:correlation}

To assess whether automated XAI metrics align with human judgments of explanation quality, we computed Pearson correlations between seven CF metrics and the five rating dimensions as well as the CQS, separately for each dataset.

\begin{figure}[t]
    \centering
    \includegraphics[width=\textwidth]{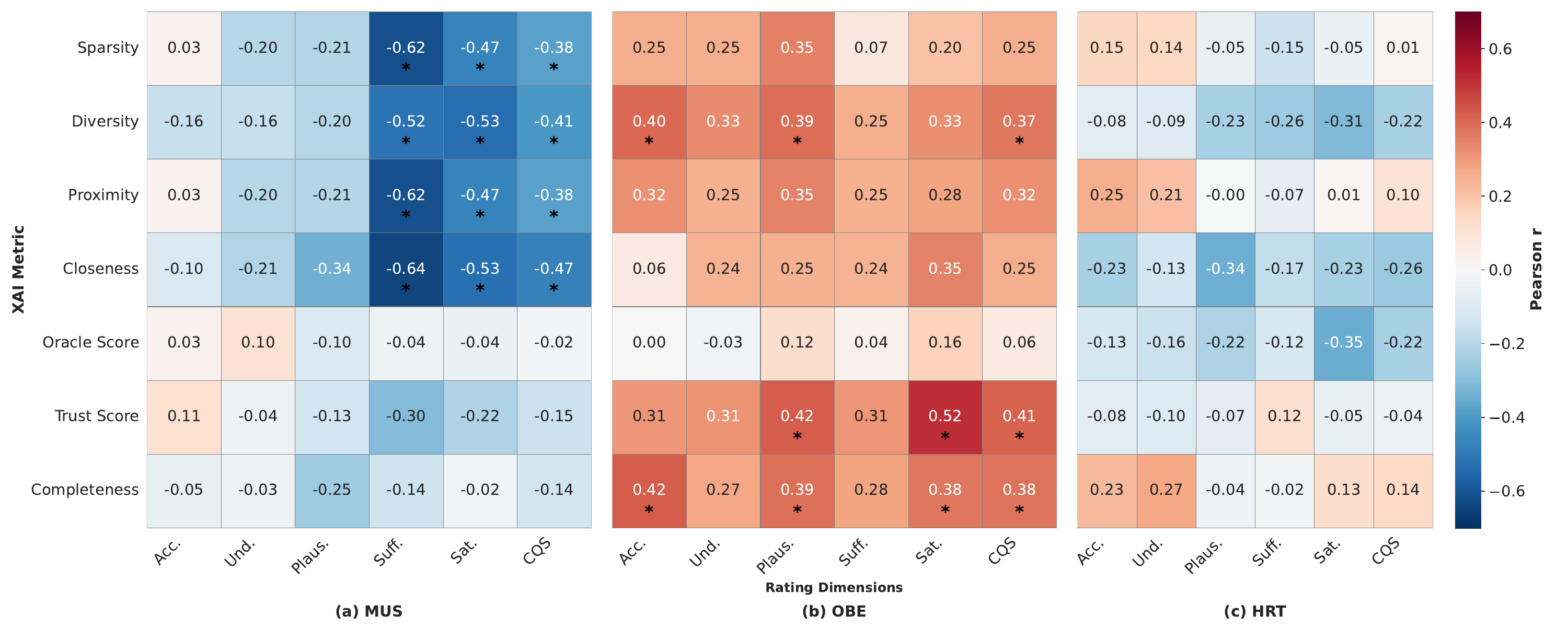}
    \caption{Metric--rating correlations including the CQS for (a) MUS, (b) OBE, (c) HRT datasets. Significant correlations ($p<0.05$) are marked with an asterisk.}
    \label{fig:metric-rating-corr}
\end{figure}

As can be seen in Figure~\ref{fig:metric-rating-corr}, correlations are generally weak across datasets. Only \textit{trust score} shows a statistically significant association with the CQS when aggregating across all explanations ($r=0.307$, $p=0.004$), while all other metrics exhibited negligible correlations ($|r|<0.1$). This indicates that individual automated metrics show limited alignment with human judgments of CF quality.

At the dataset level, however, correlation patterns differ substantially. MUS and HRT are binary classification tasks, whereas OBE is a multi-class problem with seven target classes, a distinction that may help explain the divergent pattern across datasets described below. In the MUS dataset (Figure~\ref{fig:metric-rating-corr} (a)), several metrics --- including \textit{sparsity}, \textit{diversity}, \textit{proximity}, and \textit{closeness} --- are moderately to strongly negatively correlated with \textit{sufficiency of detail}, \textit{satisfaction}, and CQS ($r=-0.38$ to $-0.64$). This suggests that users in this domain prefer CFs involving fewer and smaller changes.

In contrast, results for the OBE dataset (Figure~\ref{fig:metric-rating-corr} (b)) show mostly positive correlations. Here, \textit{diversity}, \textit{trust score}, and \textit{completeness} are positively correlated with multiple rating dimensions, including \textit{plausibility}, \textit{satisfaction}, and the CQS ($r=0.37$ to $0.52$). This indicates a preference for more comprehensive or information-rich explanations in this domain.

Finally, for the HRT dataset (Figure~\ref{fig:metric-rating-corr} (c)), correlations are uniformly weak and non-significant across all metrics and rating dimensions, with mixed directions and small effect sizes. No consistent relationship between automated metrics and perceived explanation quality emerges for this dataset.

Overall, both the direction and magnitude of metric--rating correlations vary substantially across datasets (mean cross-dataset standard deviation of correlation coefficients: $0.31$). No single metric, nor any consistent metric combination, reliably proxies human judgments across domains. These results indicate that relationships between automated CF metrics and human perception are highly dataset-specific rather than universal.

\subsection{Predictive Modeling} \label{sec:predicting}

While the correlation analysis discussed in the previous section assesses whether individual metrics align with human judgments, it does not address whether combinations of metrics can jointly approximate perceived explanation quality --- an assumption that implicitly underlies many evaluation practices in which multiple automated metrics are reported side by side. We therefore evaluate the predictive power of metric combinations using supervised learning models.

We perform an exhaustive powerset analysis over all 127 non-empty subsets of the set of seven metrics. For each subset, we train five model classes across three datasets and six target variables (five rating dimensions and the CQS): linear regression, k-nearest neighbors (kNN), Random Forest (RF), XGBoost, and generalized additive models (GAMs). Performance is evaluated using 5-fold cross-validated $R^2$, which measures generalization relative to a mean baseline.

Across all settings, predictive performance is generally poor. Linear regression models consistently fail, yielding strongly negative $R^2$ values (mean $R^2=-1.253$), indicating that linear combinations of metrics do not explain variance in human ratings. Among non-linear approaches, XGBoost underperforms even linear models (mean $R^2=-1.874$), likely due to overfitting in this low-sample regime, while GAMs frequently fail to converge. kNN yields modest improvements (mean $R^2=-0.887$), but still remains below baseline. RFs perform best overall (mean $R^2=-0.474$), yet predictive power remains limited.

\begin{figure}[t]
    \centering
    \includegraphics[width=0.9\textwidth]{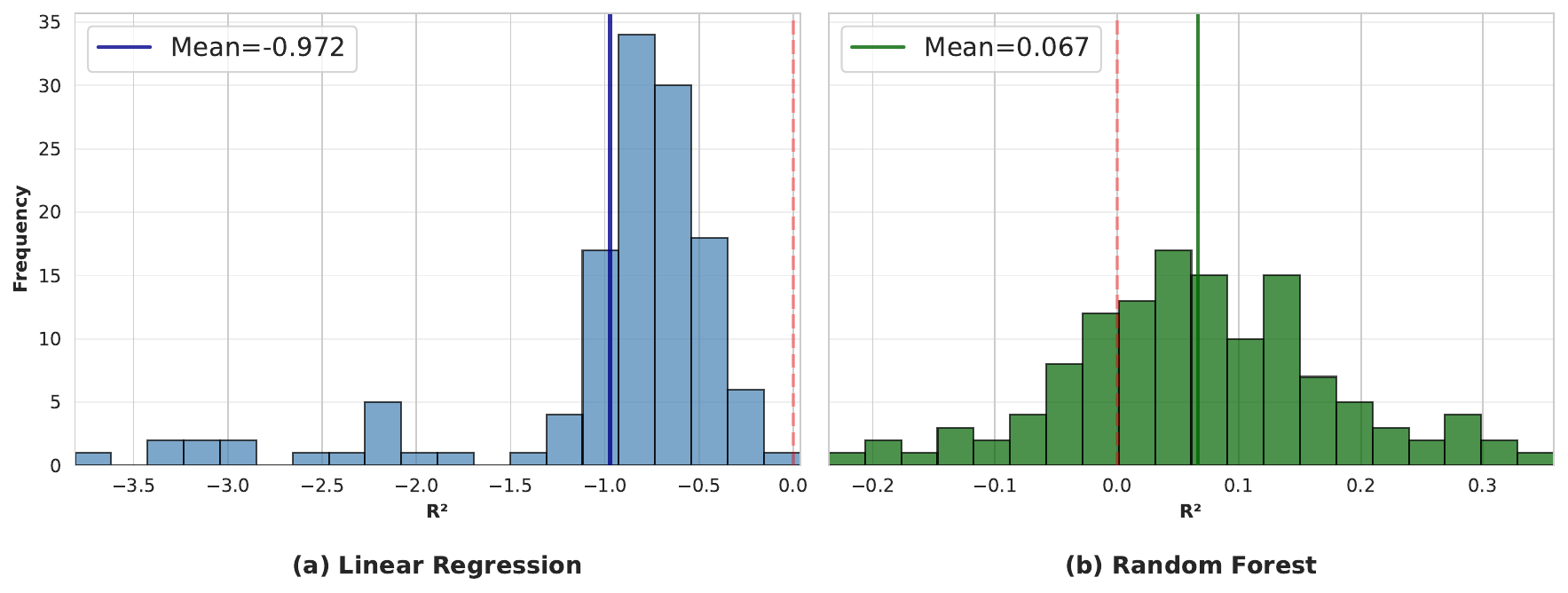}
    \caption{Histograms of $R^2$ for (a) linear regression and (b) the best performing non-linear model, RF, on the HRT dataset predicting user \textit{satisfaction} rating.}
    \label{fig:histogram_non_linear}
\end{figure}

Given the consistent pattern across datasets and target dimensions, Figure~\ref{fig:histogram_non_linear} illustrates a representative case: predicting user \textit{satisfaction} for the HRT dataset. All linear models yield negative $R^2$ values ($M=-0.972$), confirming the absence of predictive signal. In contrast, RF, the best performing non-linear model, achieves positive $R^2$ values in 95 of 127 metric combinations (range: $-0.209$ to $0.331$, $M=0.067$). While non-linearity improves performance, the resulting models still explain only a small fraction of the variance.

\begin{figure}[t]
    \centering
    \includegraphics[width=0.9\textwidth]{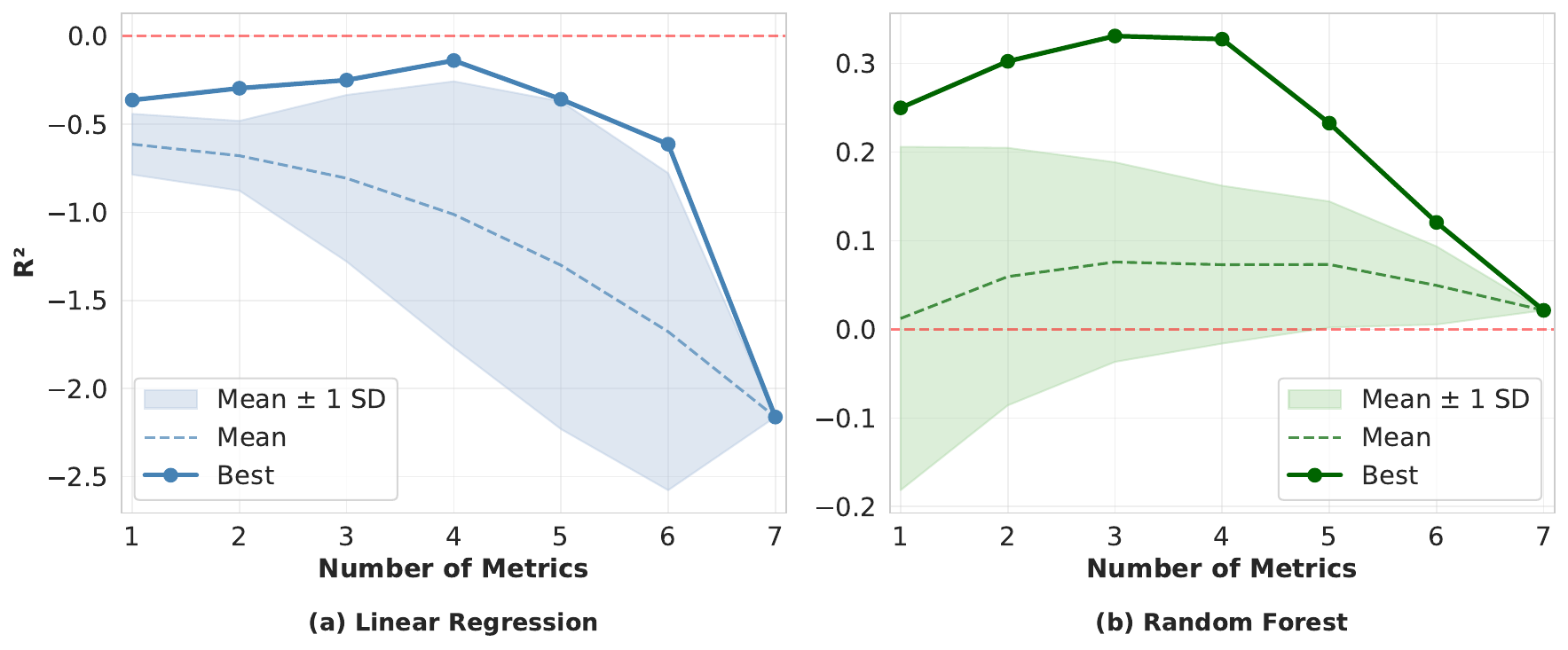}
    \caption{$R^2$ by number of included metrics for (a) linear regression and (b) RF, the best performing non-linear model, on the HRT dataset predicting user \textit{satisfaction} rating. Lines show the best and mean $R^2$ across all metric combinations at each complexity level, with shaded bands indicating $\pm 1$ SD.}
    \label{fig:r2_model_complexity}
\end{figure}

Figure~\ref{fig:r2_model_complexity} illustrates the analysis of predictive performance as a function of \textit{model complexity}, defined as the number of metrics included in a given combination. Each side panel shows the best, mean, and mean $\pm1$ SD $R^2$ values across all 127 metric combinations at each complexity level ($k=1, \ldots, 7$ used metrics).
For linear regression (Figure~\ref{fig:r2_model_complexity} (a)), $R^2$ remains strictly negative across all complexity levels. Performance improves slightly when increasing the number of metrics up to four, but deteriorates sharply as additional metrics are added, suggesting increasing noise without added explanatory value.

A similar pattern emerges for RFs (Figure~\ref{fig:r2_model_complexity} (b)). While mean $R^2$ values are positive, they remain below $0.1$ across all complexity levels, indicating weak predictive power. The best models peak at three and four metrics (maximum $R^2=0.33$ for three used metrics) and then decline monotonically as more metrics are added. Thus, even for non-linear models, increasing the number of metrics beyond a certain value ($\approx3-4$) does not improve, but degrades performance.

Figures~\ref{fig:histogram_non_linear} and~\ref{fig:r2_model_complexity} provide little evidence that existing CF metrics or their combinations can reliably predict human judgments of explanation quality. While non-linear models yield modest improvements over linear baselines, predictive performance remains weak overall, even for the best model, underscoring limitations of using current automated metrics as proxies for human evaluation.

\subsection{Discussion}

Across correlation and predictive analyses, our results consistently show that widely used automated CF evaluation metrics align only weakly with human judgments of explanation quality. Individual metric--rating correlations are generally small and highly dataset-dependent, and combining multiple metrics does not yield robust improvements in predictive performance. Even the best-per\-for\-ming non-linear models explain only a limited fraction of variance in human ratings and often degrade as more metrics are added. Together, these findings suggest that current automated CF metrics, whether considered individually or in combination, cannot serve as reliable proxies for human evaluation of explanation quality.

These results confirm and extend prior observations in human-centered XAI research. This aligns with cognitive and empirical studies showing that user \textit{satisfaction} and \textit{trust} depend on context and cannot be reduced to single criteria~\cite{byrne2019counterfactuals,warren2022features}. Similarly, work highlighting divergences between computational plausibility and psychological plausibility~\cite{kuhl2022keep}, as well as broader evidence from XAI and adjacent fields~\cite{li2023m4,sun2023privacy}, point in the same direction. Crucially, we extend prior work that models human judgments from human-rated attributes~\cite{domnich2025predicting} by showing that this structure does not readily translate to automated metrics. Taken together, our results highlight a structural gap between how CFs are currently evaluated and how they are experienced by users, underscoring the need for evaluation approaches that are more directly grounded in human judgment.

\section{Conclusion}

We examined whether widely used automated metrics for CFs reflect user-perceived explanation quality. Across datasets, rating dimensions, and analytical approaches, we find little evidence that existing metrics and their combinations align with human judgments. Metric--rating associations are weak and highly context-dependent, and increasing the number of metrics does not improve predictive performance. These findings challenge the common practice of treating automated CF metrics as reliable proxies for human evaluation and suggest that current metric-based assessments provide an incomplete picture of explanation quality. More broadly, our results underscore the need to rethink how CFs are evaluated if XAI systems are to be assessed in ways that meaningfully reflect human judgment.\\

\noindent\textbf{Limitations.}
While our study provides initial evidence on the alignment between automated CF metrics and human judgments, several aspects of the study design bound the generalizability of our findings. First, the scope is constrained in terms of the number of datasets, participants, metrics, and the single CF generation method used. Although our analysis did not reveal strong effects, increasing the number of annotators and including additional datasets could enable the detection of small or medium-sized effects that our current sample size is not powered to identify. Second, although a valid CF should in principle be independent of the generation method and our analytical approach is agnostic to how explanations are produced, our study relies on a single generation method. Evaluating CFs produced by alternative generation approaches would strengthen the generalizability of our findings. Third, while demographic information and self-reported ML experience were collected, the study does not include a domain knowledge assessment specific to the tasks associated with each dataset. Although datasets were selected for their intuitive features, varying levels of background knowledge may nonetheless influence how participants evaluate CFs. Additionally, the study relies on a general online participant pool via Prolific, meaning our findings most directly reflect lay user perception. The extent to which results generalize to domain experts remains an open question and a direction for future work.\\

\noindent\textbf{Future Work.}
Building on the limitations identified above, we see several directions for future research. First, we aim to develop automated proxy metrics that are grounded in human-centered theory and validated against user perception, with the goal of better approximating how users evaluate CFs. Second, additional empirical studies across tasks, modalities, and explanation types are needed to determine which properties users truly value and how these can be operationalized in reliable, human-aligned evaluation methods. Third, future work should investigate further contributing factors to perceived explanation quality, such as the actionability of CFs, which may influence user judgments but is not captured by current metrics.

\begin{credits}
\subsubsection{\ackname} This research was partially funded by the \textit{Deutsche Forschungsgemeinschaft}: TRR 318/1 2021 -- 438445824, the \textit{Ministry of Culture and Science of North Rhine-Westphalia} under the grant No.~NW21-059A \href{https://www.sail.nrw/}{SAIL}, and by the \textit{Research Council of Norway} through its Centre of Excellence Integreat --- The Norwegian Centre for knowledge-driven machine learning, project number 332645.

\subsubsection{\discintname} The authors have no competing interests to declare that are relevant to the content of this article.
\end{credits}
%
%
%
\bibliographystyle{splncs04}
\bibliography{references}

\end{document}